\documentclass[11pt]{article}
\usepackage{coling2016}
\usepackage{times}
\usepackage{url}
\usepackage{latexsym}
	
\usepackage{amsmath}
\usepackage{graphicx}
\usepackage{epstopdf}
\usepackage{subfigure}
\usepackage{booktabs}
\usepackage{threeparttable}
\usepackage[labelformat=empty,labelsep=none]{caption}
\usepackage[justification=justified,singlelinecheck=off]{caption}

\title{Evaluation method of word embedding by roots and affixes}

\author{KeBin Peng \\
  School of Computer Science and Engineering  \\
  Beihang University \\
  Beijing 100191 \\
  {\tt kebinpeng@act.buaa.edu.cn}}

\date{}

\begin{document}
\maketitle
\begin{abstract}
  Word embedding has been shown to be remarkably effective in a lot of Natural Language Processing tasks. However, existing models still have a couple of limitations in interpreting the dimensions of word vector. In this paper, we provide a new approach---roots and affixes model(RAAM)---to interpret it from the intrinsic structures of natural language. Also it can be used as an evaluation measure of the quality of word embedding. We introduce the information entropy into our model and divide the dimensions into two categories, just like roots and affixes in lexical semantics. Then considering each category as a whole rather than individually. We experimented with English Wikipedia corpus. Our result show that there is a negative linear relation between the two attributes and a high positive correlation between our model and downstream semantic evaluation tasks.
\end{abstract}

%

\section{Introduction}
\label{intro}

%
%
\blfootnote{
    %
    %
    \hspace{-0.65cm}  
     This work is licensed under a Creative Commons Attribution 4.0 International Licence. Licence
     details: http://creativecommons.org/licenses/by/4.0/
    %
    %
    %
    %
}

Distributed representation of word has become increasingly popular currently. It can be utilized to various downstream NLP tasks. For example, bilingual word embedding for phrase-based machine translation \cite{zou2013bilingual}, enhancing the coverage of POS tagging \cite{huang2014learning}. Success of word embedding inspires us to go deep into the structure of distributional representations. Existing works such as CBOW and skip-gram in the toolbox word2vec \cite{mikolov2013efficient} using large unlabeled corpora to train the model. However, every single dimension in the vector has no obvious meaning. We are not yet clear why it is efficient. Hence, interpreting the meaning of the dimension may be essential to answer these questions.

  There are some notable interpretation embedding model includes Subspace Alignment \cite{tsvetkov2015evaluation}, Non-distributional Word Vector Representations \cite{faruqui2015non}. Non-Negative Sparse Embedding \cite{murphy2012learning}, Online Learning of Interpretable Word Embedding \cite{luo2015online} and so on. They use existing tools, for example, WordNet, to allocate the dimension with a certain meaning. Although these models have revealed some dimensions in vector do have clearly meaning, it turns out that they are not suited for high dimension vector because the length of WordNet is limited but the vector can be very long. Meanwhile, they do not consider the sequence of allocation.

In this paper, we propose a new efficient model named roots and affixes model(RAAM), to clarify the meaning of the dimensions in a word vector from the intrinsic structures of natural language---word and sentence. On the one hand, word and sentence are the natural structure in text, we assume this structure should be incarnated in word vector. On the other hand, we introduce the information entropy into our model and define two attributes to a dimension which named: word entropy and sentence entropy. Through the two attributes, we can divide the dimensions in word vector into two classes at different level: word level and sentence level, just like roots and affixes in vocabulary. Different roots and affixes represent different semantic meaning. Analogously, the different level of word vector represents different semantic meaning in word vector. So, we believe these two attributes represent two aspects of word vector.

Compared with previous works, we have made three main contributions:
(1) we promote a new conception named roots and affixes model(RAAM) which are inspired by the entropy and lexical semantics. The conception can explain RAAM clearly
(2) we define two new attributes for the dimensions in word vector which can reflect two distinct levels of semantic.
(3) we discuss the interpretable embedding from a new point. Our model gives a very obvious hierarchies relationship among and sentence. These hierarchies are the natural structure, not man-made.

The rest of this paper is organized as follows: In the next section, we summarize some previous works. In section 3, we present our model formally. In section 4, we report experimental results. Finally, we conclude in section 5.

\section{Related works}

In this section, we discuss some related works. Several interpretable embedding models have been developed to explain the meaning of the dimensions in a word vector. It can be partitioned into two kinds of methods. First, In \cite{tsvetkov2015evaluation} \cite{faruqui2015non}, word level information from linguistic resources such as WordNet is extracted to construct word vectors. It dimensions can be integer or decimal. But, this representation cannot give meaning to all the dimensions because the length of the word vector is much longer than the attributes of a word in WordNet. Second, In\cite{murphy2012learning}, distributional models which apply matrix factorization are introduced. In this model, word representations learned by Non-Negative Sparse Embedding is sparse, effective, and interpretable. The second method utilizes various object functions to train the word vector. Such as cite{luoonline}\cite{fyshe2014interpretable}\cite{lin2007projected}. These models apply some new object functions such as projected gradient descent and no negative constraints for optimization. Nonetheless, Only few dimensions can be interpreted clearly. In addition, all the approach mentioned above do not show the relationship between words and sentences. Meanwhile, if we look the word vector as a kind of encoding, a dimension may be meaningless unless get them together. Hence, we are motivated to consider the dimensions as a whole and try to reveal the relationship between words and sentences.

\section{Word Vector Dimension Interpret Model}
\label{sect:pdf}

In this section, we introduce the basic structure of our model. Our model comprises 3 phases. In the first phase,
We introduce the information entropy into our model and define two attributes for a dimension in the word vector: word entropy and sentence entropy. In the second phase, we calculate the two different attributes according to the formulations which defined in the first phrase. In the third phrase, we separate the word vector dimensions into two parts which we named word level and sentence level using the attributes stated above. In this section, we formally describe the model, which we call RAAM.

The RAAM's underlying hypothesis is that ith dimension in word vector is contributing to ith dimension in sentence vector or paragraph vector. It is motivated by two reasons: first, sentence is a natural structure in text rather than manual work. We guess the semantic meaning could be reflected in these natural structure. Second, explaining a single dimension is problematic. Because the vector dimension can be 300 or more. It is intricate to give every dimension a clearly meaning. So we attempt to get the dimension together and give an interpret them in the natural structure that a document has.

Formally, there is a word (sentence) vector v=\{ $p_{1}$ \dots $p_{i}$ \dots $p_{l}$  \}. We look the $p_{i}$ as a random variables. Also, we have a matrix

%
%

\begin{equation}\label{Eq:matrix1}
\bordermatrix{%
          & dimention_1  &\cdots   &dimention_1\cr
word vector_1    & p_{11}       &\cdots   & p_{1m}\cr
\vdots    & \vdots       &\vdots   &\vdots\cr
word vector_m    & p_{n1}       &\cdots   &p_{nm}
},
\end{equation}

$p_{nm}$ is as same as $p_{i}$. And then, we use following two formulas to calculate the entropy of i dimension to the word level.
\begin{equation}
p_{j} = exp[ -\frac { ( p_{i}-\mu )^{2} } {2\sigma^{2}} ]
\end{equation}

\begin{equation}
 E^{w}_{k}(i) = \sum_{j=1}^l p_{j}logp_{j}
\end{equation}

$E^{w}_{k}(i)$ Means the entropy of i dimension of k word to the word level.Second, we calculate the entropy of the sentence level. We have a sentenced vector v=\{ $p_{1}$ \dots $p_{i}$ \dots $p_{t}$  \}. So, according to our model, the entropy to the sentence vector is calculated by following formula:

\begin{equation}
E^{s}_{m}(i) = \sum_{j=1}^t p_{j}logp_{j}
\end{equation}

Finally, we concern the interpretation as a process which could decrease the information entropy. When we look at a word in a sentence, we get some information. That means the information entropy has been lowered. Further more, as information in word and sentence contains each other, we use mutual information which defines as

\begin{equation}
I(X;Y) = \sum_{x\in X}\sum_{y\in Y} p_{xy}(i,j)log \frac{p_{xy}(i,j)}{p(x)p(y)}
\end{equation}

The number of $I(x;y)$ means how much information does y have when we get x. x or y can be word, sentence, and paragraph. Information that y contains is inversely proportion to the number of $I(x_i;y_i)$.

we calculate the entropy of the i dimension to the sentence level. $p_{w}(i)$ is the element of the matrix. Polk is the element in the sentence vector. $p_{k}$ is the element in the sentence vector.

\begin{equation}
P_{xy}(i,j) = \frac{1}{\sqrt{2\pi}\sigma_0\sigma_1}exp(-\frac{(p_w(i)-\mu_0)^2}{2\sigma_0^2})exp(-\frac{(p_k(i)-\mu_1)^2}{2\sigma_1^2})
\end{equation}

If we write exactly, we can get following formula:

\begin{equation}
I(X;Y) = \sum^l_{i=1}\sum^t_{j=1}P_xy(i,j)log(\frac{P_{xy}(i,j)}{P_j|E^{s}_{m}(i)|})
\end{equation}

So, for every dimension, it has two attributes $E_m^s(i)$ and $E_k^w(i)$. This two attributes is a measure to a dimension. Through the experiment, we find the that there is a negative linear relationship between the two attributes. Thus, we can decide which attribute is more important and divide the whole dimension into two level: word level and sentenced level. For a dimension in word vector, for example, if the $E_m^s(i)$ of this dimension is larger than$E_k^w(i)$, we can get a conclusion that this dimension of the word contributes more information to the sentence which the word belongs to than word itself. We can get all this kind of dimension into a class. we believe that class interpret the meaning of the whole sentence.
The finally score is

\begin{equation}
E_T =\sum_{i=1}^l max(E_m^s(i),E_k^w(i))
\end{equation}
\section{Experiment}
\label{ssec:Experiment}

We conducted sufficient experiment on real-world dataset to verify our model. All the codes were implemented in MATLAB. We select some state-of-the-art word vector models.

\subsection{Word vector models}

We chose some state-of-art word vector models and compare those models with our RAAM model. We used English Wikipedia dataset of RAAMX. The vector dimension is 50, 100, 150, 200, 250, 300, 350, 400 500.\\

\noindent\textbf{Skip-Gram(SG) and CBOW}.    Word2Vec tool \cite{mikolov2013efficient}  \cite{mikolov2010recurrent}  \cite{le2014distributed} is very popular in nature language processing and effectively. This model uses Huffman code to represent the word. Then considering it as input to log classifier. The word is predicted within a given context window.\\

\noindent\textbf{Glove}.    GloVe is an unsupervised learning algorithm for obtaining vector representations for words. Training is performed on aggregated global word-word co-occurrence statistics from a corpus, and the resulting representations showcase interesting linear substructures of the word vector space\cite{pennington2014glove}.\\

\noindent\textbf{Glove+WN, Glove+PPDB}.     We used the WordNet(WN), paraphrase database(PPDB) \cite{ganitkevitch2013ppdb}  and  \cite{faruqui2014retrofitting} to enrich the semantic information of the word vector.

\subsection{Evaluation Benchmarks}

We compare our RAAMX model with some standard semantic tasks. We will introduce these tasks briefly.\\

\noindent\textbf{Word Similarity}.\ \ There are three different kinds of benchmarks: \textbf{WS-353, MEN, SimLex-999}.
The \textbf{WS-353} contains two sets of English word pairs along with human-assigned similarity judgements. The collection can be utilized to train and/or test computer algorithms implementing semantic similarity measures\cite{finkelstein2001placing}. The \textbf{MEN} contains two sets of English word pairs (one for training and one for testing) together with human-assigned similarity judgments. The collection can be used to train and/or test computer algorithms implementing semantic similarity and relatedness measures\cite{bruni2014multimodal}. \textbf{SimLex-999} is a gold standard resource for the evaluation of models that learn the meaning of words and concepts\cite{hill2015simlex}. It can overcome the shortcomings of \textbf{WS-353} and contains 999 pairs of adjectives, nouns and verbs\cite{tsvetkov2015evaluation}. We computer word similarity by cosine similarity.

\noindent\textbf{Semantic Task}. Our semantic task is sentiment analysis task(Senti). Senti\cite{socher2013recursive} is a classification task which between positive and negative movie reviews.

\section{Results}

At the beginning, we display the relationship between the two hierarchies: word and sentence.

\begin{figure}[!htbp]
\begin{minipage}[t]{0.3\linewidth}
\centering
\includegraphics[width=2in, height=2.0in]{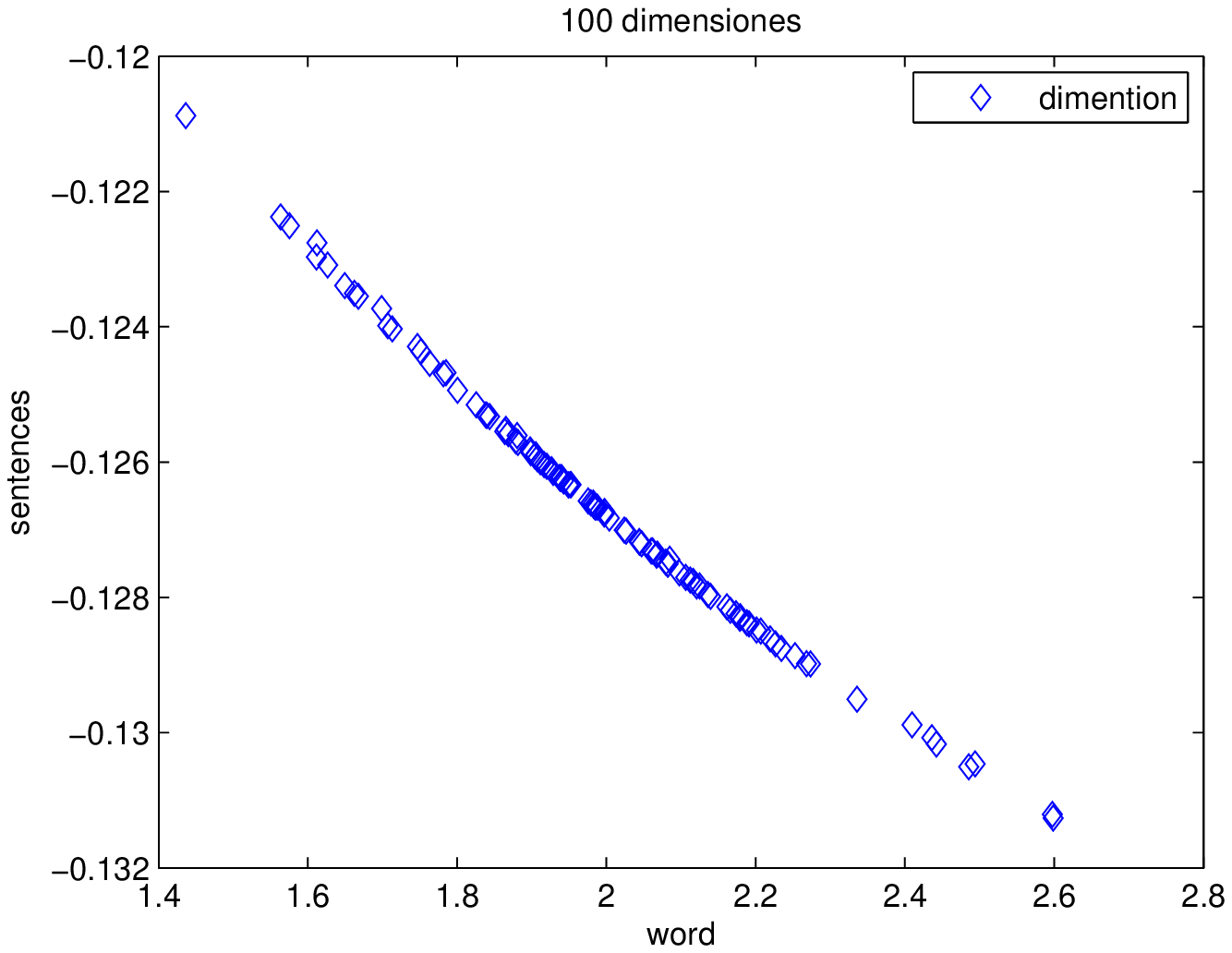}

\label{fig:side:a}
\end{minipage}%
\begin{minipage}[t]{0.3\linewidth}
\centering
\includegraphics[width=2in, height=2.0in]{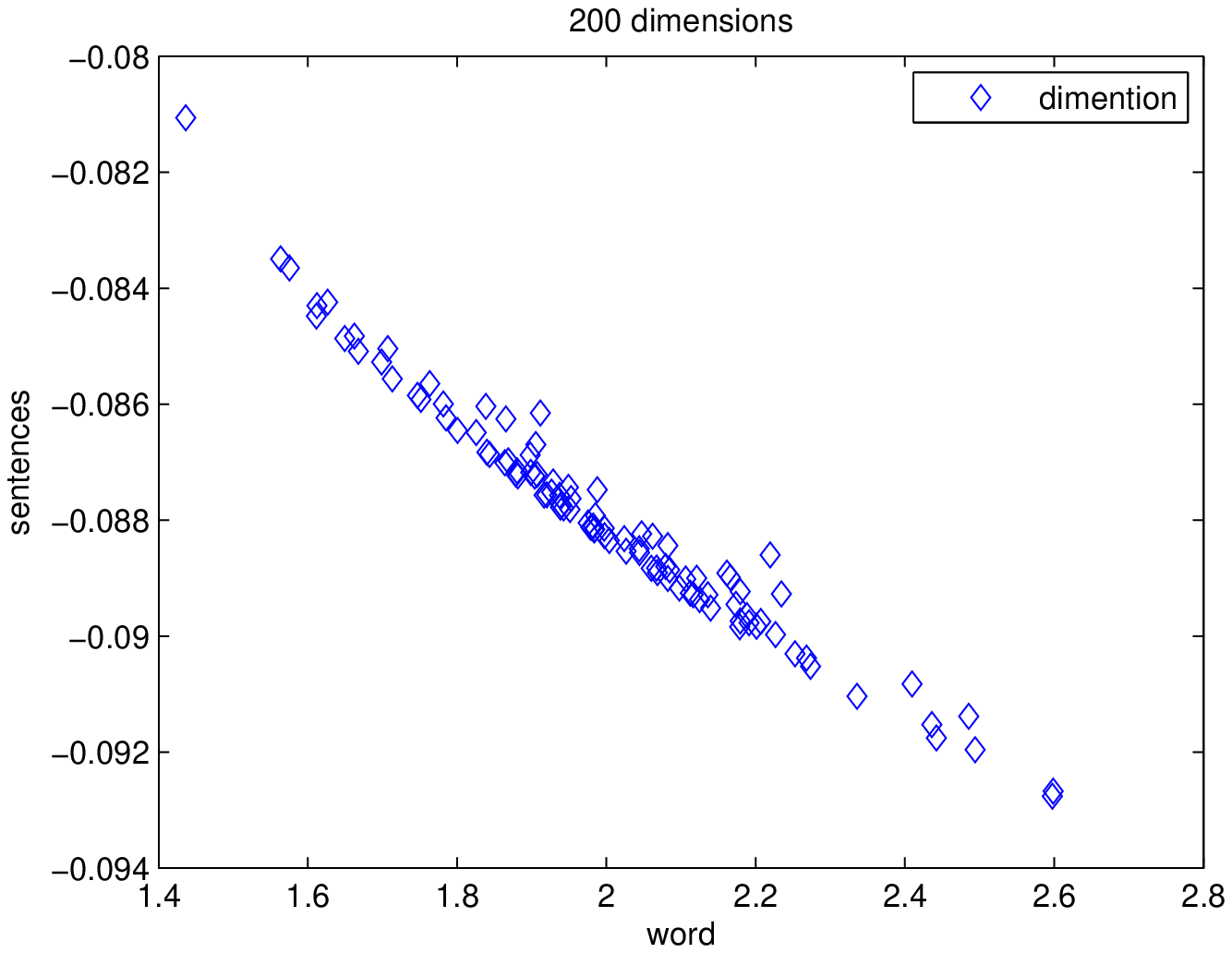}
\end{minipage}
\begin{minipage}[t]{0.3\linewidth}
\centering
\includegraphics[width=2in, height=2.0in]{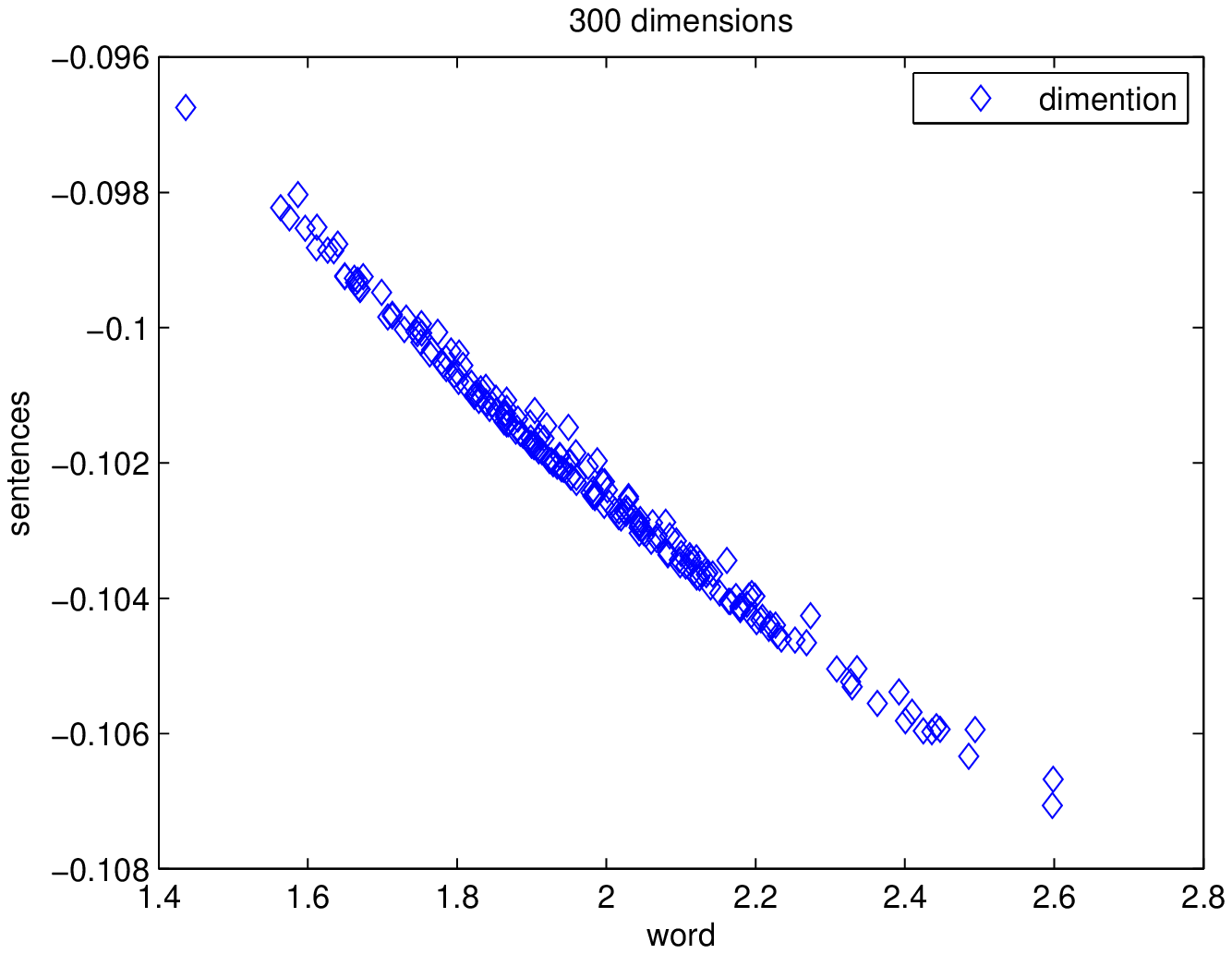}
\label{}
\end{minipage}

\caption{Figure 1: The relationship between word level and sentence level in different dimension.}
\end{figure}

From the figure1, 2, 3, we can draw a conclusion the there is a negative linear relationship between the two levels. And the linear relationship is not changed among the different of parameter. This conclusion proves our hypothesis that there is difference between the two hierarchies for one dimension.

Later, according to \cite{tsvetkov2015evaluation},  we compute the Pearon¡¯s correlation coefficient r between RAAMX¡¯s scores and other word vector models. Our purpose is checking the correctness between these models. First, we use the task named Senti to make comparison. We set the vector dimensions in 300.

\begin{table}[!htbp]

\centering
\caption{}
\begin{tabular}{ccc}

\toprule
Model&my scour&senti\\
\midrule
CBOW&200.1667&90\\
SG&199.3584&80.5\\
GloVe&180.8564&79.4\\
GloVe+WN&178.7853&79.6\\
GloVe+PPDB&176.1831&79.7\\
LSA&169.1976&76.9\\
LSA+WN&165.4816&77.5\\
lsa+PPDB&164.4703&77.3\\
\midrule
Pearon's correlation&\multicolumn{2}{c}{0.790310178}\\
\bottomrule

\end{tabular}
\begin{tablenotes}
    \item [1]Table 1: scores of RAAM, extrinsic task, and Senti task. The word vector using different models.

   \end{tablenotes}

\end{table}


%
%
%
%

As we display on table 1, the Pearson¡¯s correlation between the RAAMX scores and other models scores is r=0.7903.
Second, we show the correlation between two different train ways and RAAM scores in different dataset. From the table, we can see there is a high positive correlation between RAAM scores and two different train ways.

Second, we show the correlation between two different train ways and RAAM scores in different dataset. From the table 2 we can see there is high positive correlation between RAAM scores and two different train ways.

\begin{table}[!htbp]
\centering
\caption{}
 \begin{tabular}{cccc}
\toprule
r&WS-353&MEN&SimLex\\
\midrule
CBOW and RAAM&0.4753&0.6231&0.7014\\
SG and RAAM&0.5869&0.6721&0.6451\\
\bottomrule
\end{tabular}
\begin{tablenotes}
    \item [1] Table 2: The correlation between CBOW, SG, and RAAM in different corpuses.

   \end{tablenotes}
\end{table}

Next, we show the correlation of RAAM with the different task and different word vector models with different dimensionality:

\begin{table}[!htbp]
\centering
\caption{ }
\begin{tabular}{ccccc}
\toprule
&WS-353&MEN&SimLex&Senti\\
\midrule
r(100)&0.346621&0.536587&0.606502&0.641462\\
r(200)&0.382454&0.509717&0.633696&0.623471\\
r(300)&0.435112&0.528295&0.659765&0.670678\\
\bottomrule
\end{tabular}

\begin{tablenotes}
    \item [1] Table 3:  RAAM with the different task and different word vector models with
different dimensionality.

   \end{tablenotes}
\end{table}

Table 3 show that RAAM obtains a high positive correlation with downstream tasks.

Finally, we show the correlation between RAAM score and Senti in different dimension in table 4. From the table we can see, the highest correlation is in the 400dimensions. After this point, the correlation starts to fall
%
%

\begin{table}[!htbp]
\centering
\caption{}
\begin{tabular}{cc}
\toprule
Dimensions & R  \\
\midrule
50&0.276424\\
100&0.306564\\
150&0.415664\\
200&0.462858\\
250&0.583681\\
300&0.699617\\
400&0.775584\\
500&0.620081\\
1000&0.543019\\
\bottomrule
\end{tabular}

\begin{tablenotes}
    \item [1] Table 4: correlation between RAAM score and Senti in different dimension.

   \end{tablenotes}

\end{table}

To summarize, we observe a high positive correlation between RAAM and the downstream tasks, and across discrete models with vectors of different dimensionalities.

\section{Conclusion}

In this paper, we visit the problem of the interpretable embedding and propose a new approach to interpret the meaning of the dimensions in a word vector. We design two attributes of a dimension in a word vector. According to the two attributes, we aggregate dimensions into two classes. We have experimented on several dataset. Our result suggests that the difference two class captures two levels of semantics of a word. We believe our approach delivers valuable information and provide a good way that interpret the meaning of a dimension for future research.

\section*{Acknowledgement}
We want to say thank you to all the anonymous reviewers for constructive feedback.

\bibliographystyle{acl}
\bibliography{Bibtex}

%
%
%
%
%
%

\end{document}